\documentclass[10pt,twocolumn,letterpaper]{article}

\usepackage{wacv}
\usepackage{times}
\usepackage{epsfig}
\usepackage{graphicx}
\usepackage{amsmath}
\usepackage{amssymb}
\usepackage{float}
\usepackage{multicol}
\usepackage{multirow}
\usepackage{capt-of,etoolbox}
\usepackage{booktabs}

\def\wacvPaperID{****} 

\wacvfinalcopy 
\ifwacvfinal
\usepackage[breaklinks=true,bookmarks=false]{hyperref}
\else
\usepackage[pagebackref=true,breaklinks=true,colorlinks,bookmarks=false]{hyperref}
\fi

\begin{document}

\title{DG-Labeler and DGL-MOTS Dataset: \\Boost the Autonomous Driving Perception}

\author{Yiming Cui\footnotemark[1] \\
University of Florida\\
Gainesville, FL 32611\\
{\tt\small cuiyiming@ufl.edu}
\and
Zhiwen Cao\footnotemark[1]\\
Purdue University\\
West Lafayette, IN 47907\\
{\tt\small  cao270@purdue.edu}
\and
Yixin Xie\\
The University of Texas at El Paso\\
El Paso, TX 79968\\
{\tt\small   yxie4@miners.utep.edu}
\and
\hspace{-1em}
Xingyu Jiang \\
\hspace{-1em}
Purdue University\\
\hspace{-1em}
West Lafayette, IN 47907\\
{\tt\small\hspace{-1em}  jiang718@purdue.edu}
\and
Feng Tao \\
The University of Texas at San Antonio\\
San Antonio, TX 78249\\
{\tt\small   feng.tao@my.utsa.edu}
\and
Yingjie Victor Chen \\
Purdue University\\
West Lafayette, IN 47907\\
{\tt\small   victorchen@purdue.edu}
\and
Lin Li \\
The University of Texas at El Paso\\
El Paso, TX 79968\\
{\tt\small  lli5@utep.edu}
\and
Dongfang Liu\footnotemark[2] \\
Rochester Institute of Technology\\
Rochester, NY 14623\\
{\tt\small  dongfang.liu@rit.edu}
\and
}



\twocolumn[{%
\begin{center}
    \maketitle
    \includegraphics[width=17cm]{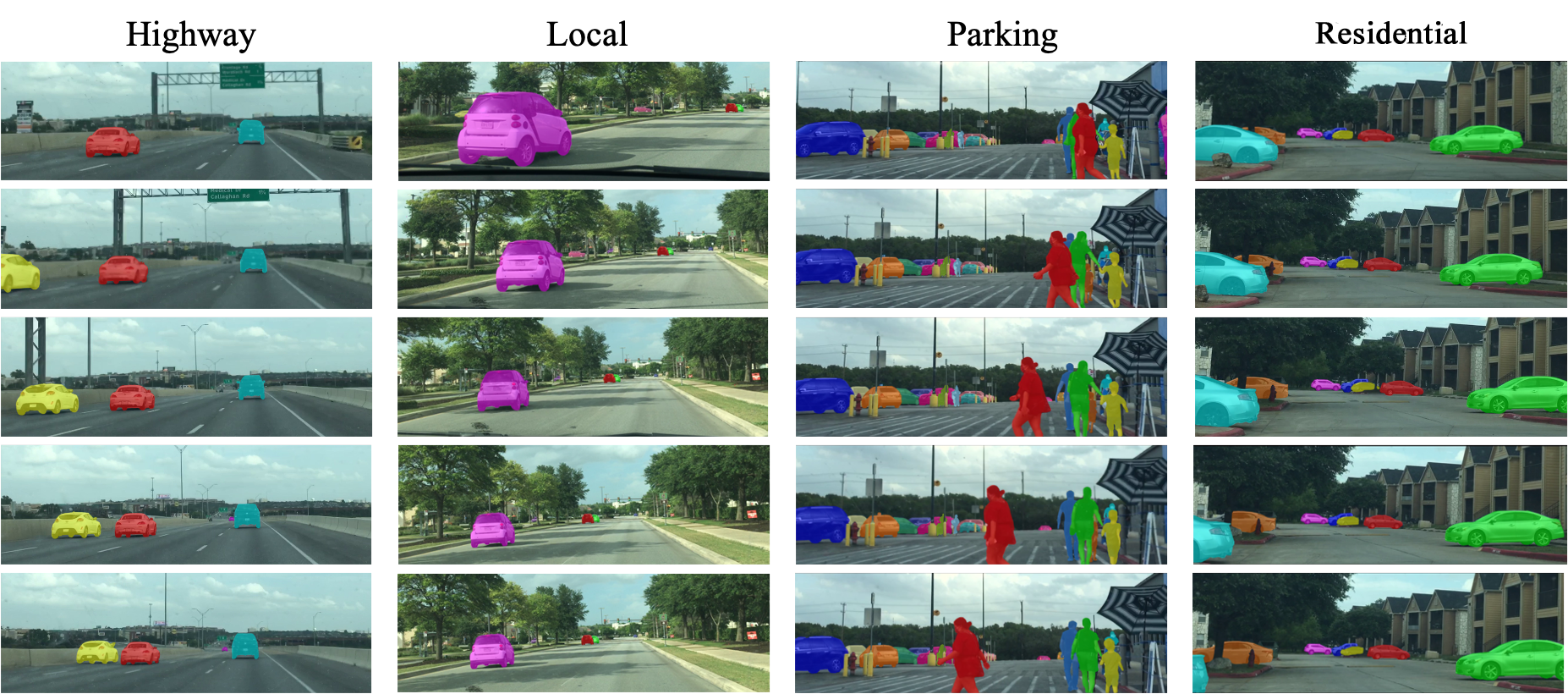}
    \captionof{figure}{A showcase of the DGL-MOTS dataset. We collect data based on different driving scenarios and organize training data based on different settings in terms of the highway, local, parking, and residential. } 
\end{center}%
}]

\renewcommand*{\thefootnote}{\fnsymbol{footnote}}
\footnotetext[1]{are equal contributions.}

\footnotetext[2]{is corresponding author.}

\renewcommand*{\thefootnote}{\arabic{footnote}}


\begin{abstract}
Multi-object tracking and segmentation (MOTS) is a critical task for autonomous driving applications. The existing MOTS studies face two critical challenges: 1) the published datasets inadequately capture the real-world complexity for network training to address various driving settings; 2) the working pipeline annotation tool is under-studied in the literature to improve the quality of MOTS learning examples. In this work, we introduce the DG-Labeler and DGL-MOTS dataset to facilitate the training data annotation for the MOTS task and accordingly improve network training accuracy and efficiency. DG-Labeler uses the novel Depth-Granularity Module to depict the instance spatial relations and produce fine-grained instance masks. Annotated by DG-Labeler, our DGL-MOTS dataset exceeds the prior effort (i.e., KITTI MOTS and BDD100K) in data diversity, annotation quality, and temporal representations. Results on extensive cross-dataset evaluations indicate significant performance improvements for several state-of-the-art methods trained on our DGL-MOTS dataset. We believe our DGL-MOTS Dataset and DG-Labeler hold the valuable potential to boost the visual perception of future transportation. Our dataset and code are available here\footnote{https://goodproj13.github.io/DGL-MOTS/}.
\end{abstract}


\section{Introduction}
A major contributing factor behind recent success in deep learning is the availability of large-scale annotated datasets \cite{russakovsky2015imagenet}. Despite the existing performance gap to humans, deep-learning-based computer vision methods have become essential advancement of real-world systems. A particularly challenging and emerging application is autonomous driving, which requires system performance with extreme reliability~\cite{liu2020video, 9206716, Cui_2021_ICCV, 9411961}. However, leveraging the power of deep learning for autonomous driving is nontrivial, due to the lack of datasets.

\indent Consequently, significant research efforts have been invested into autonomous driving datasets such as KITTI~\cite{geiger2013vision}, Cityscapes~\cite{cordts2015cityscapes}, and BDD100K~\cite{yu2020bdd100k} datasets, which serve as the driving force to the development of visual technologies for understanding complex traffic scenes and driving scenarios. As the computer vision community has made impressive advances in increasingly difficult tasks ($i.e.,$ object detection, instance segmentation, and multi-object tracking) in recent years, a new task  named  multi-object  tracking  and segmentation  (MOTS) is proposed in order to consider detection, segmentation and tracking together as interconnected problems~\cite{voigtlaender2019mots}. Consequently, the KITTI MOTS dataset~\cite{voigtlaender2019mots} is introduced to assess  the proposed  visual task.
\begin{figure*}[!ht]
	\centering
	\includegraphics[width=16.6cm]{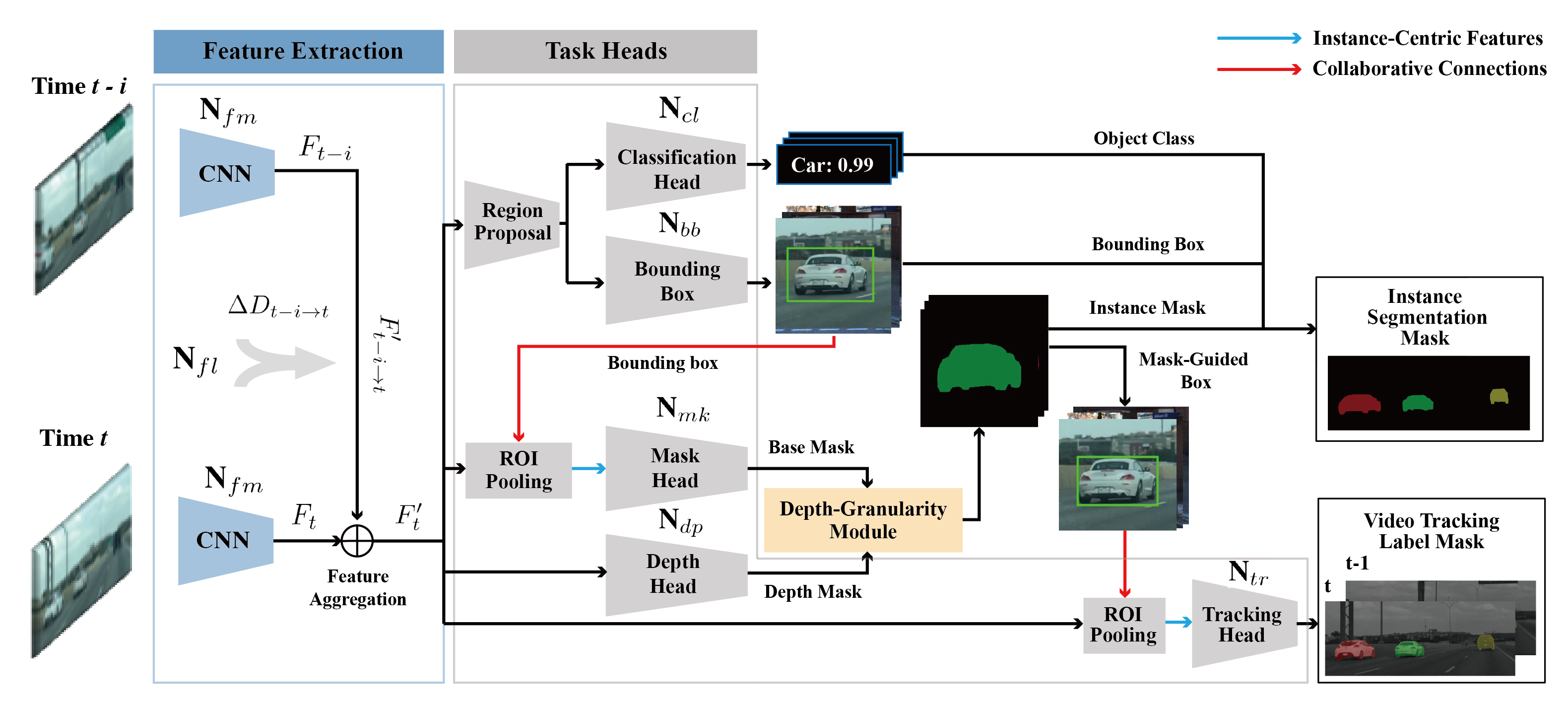}
	\caption{Illustration of DG-Labeler architecture. We craft our DG-Labeler on TrackR-CNN. In the feature extraction phase, we replace the original heavy 3D convolution with a more efficient flow network to increase the feature representation. In the task phase, we devise the collaborative connections to propagate information across each task so the upper-level task heads (a.k.a. the mask and tracking head) can perform accurate and efficient predictions on instance-centric features. Moreover, we propose a depth-granularity module, which greatly improves the segmentation behavior of our DG-Labeler.}
	\label{Frame}
\end{figure*}
\indent Although the existing MOTS datasets~\cite{voigtlaender2019mots,yu2020bdd100k} fills the gap of data shortage for MOTS task, they are limited in three significant drawbacks in the training data: \textit{1) including no challenging cases (i.e., motion blur or defocus) to address a general driving setting; 2) focusing on local roads of inner cities and thus lack diversity.} These limits may cause problems in training as the data does not fully capture the real-world traffic complexity. In addition, \textit{the annotation for MOTS data is highly labor-intensive as it requires the pixel-level mask as well as the temporal tracking label across frames.} In order to produce the MOTS data, \cite{voigtlaender2019mots} uses a refinement network to generate an initial segmentation mask followed by human corrections. Afterward, tracking labels are created by delineating the temporal coherence based on instance masks across video frames~\cite{voigtlaender2019mots}. \\
\indent To this end, we, therefore, propose the DGL-MOTS dataset to maximize synergies tailored for the MTOS task in autonomous driving. In order to improve the annotation quality and reduce the annotation cost, we devise DG-Labeler to produce fine-grained instance masks. Concretely, our work delivers the following contributions:
\begin{itemize}
    \item We create a \textbf{D}epth-\textbf{G}ranularity \textbf{L}abeled \textbf{MOTS}, thus the name of our dataset, \textbf{DGL-MOTS} (Figure~\ref{Frame}). Compared to the  KITTI  MOTS and BDD100K, DGL-MOTS significantly exceeds the previous effort in terms of annotation quality, data diversity, and temporal representation, which boosts the training accuracy and efficiency.
    \item We perform cross-dataset evaluations. Extensive experiment results indicate the benefits of our datasets. Networks trained on our dataset outperform their counterparts (with the same architecture) trained on KITTI MOTS~\cite{voigtlaender2019mots} and BDD100K~\cite{yu2020bdd100k} on the same test set. Also, improvement for 
    networks trained on our dataset is reached with less training schedule (Table~\ref{trainS}). 
    \item We propose an end-to-end annotator named DG-Labeler (Figure~\ref{Frame}), whose architecture includes a novel depth-granularity module to model the spatial relation of  instances and assist to produce fine-grained instance mask. With limited correction iterations, DG-Labeler can generate high-quality MOTS annotation.
    \item DG-Labeler leverages the depth information to depict the instance spatial relation and  retain  finer details at the instance boundary (Figure~\ref{DGM}). On both KITTI MOTS and DGL-MOTS datasets, DG-Labeler outperforms TrackR-CNN~\cite{voigtlaender2019mots} in accuracy by a significant margin. For its simplicity, we hope DG-Labeler can also serve as a new strong baseline for the MOTS task. 
\end{itemize}

\section{Related Work}
This section summarizes the related datasets for autonomous driving, multi-object tracking and segmentation as well as annotation methods for dataset creation. \\
\indent \textbf{MOTS dataset.} The multi-object tracking (MOT) is a critical task for autonomous driving, as it needs to perform object detection as well as object tracking in a video. 
A large array of datasets have been created focusing on driving scenarios, for example, KITTI tracking~\cite{geiger2012we}, MOTChallenge~\cite{milan2016mot16}, UA-DETRAC~\cite{wen2015ua}, PathTrack~\cite{manen2017pathtrack}, and PoseTrack~\cite{andriluka2018posetrack}. None of these datasets provide segmentation masks for the annotated objects and thus do not depict pixel-level representations and  complex interactions like MOTS data. More progressive datasets come from
Cityscapes~\cite{cordts2015cityscapes}, ApolloScape~\cite{huang2018apolloscape}, BDD100K~\cite{yu2020bdd100k}, and KITTI MOTS dataset~\cite{voigtlaender2019mots} which provide instance segmentation data for autonomous driving. However, Cityscapes only provides instance annotations for a small subset (i.e., 5,000 images) while ApolloScape offers no  temporal object descriptions over time. Thus, the two datasets cannot be utilized for joint training of MOTS algorithms. In contrast, KITTI MOTS~\cite{voigtlaender2019mots} is the first public dataset which fills the gap of data shortage for the MOTS task but it only includes a few thousand learning data for training; to date, BDD100K has the largest data scale from intensive sequential frames which are redundant for training. Compared to the aforementioned two datasets, our DGL-MOTS dataset includes more diverse data and fine-grained annotations.\\
\indent \textbf{Multi-object tracking and segmentation.}
The majority of MOTS methods~\cite{bertasius2020classifying,wang2020towards,zhou2020tracking,kim2020video,lin2020video,yang2019video,mohamed2020instancemotseg} intuitively extend from Mask R-CNN. Although the extension paradigm is simple, it encounters several performance bottlenecks: 1) feature sharing across each task is insufficient for joint optimization; 2) the mask head struggles to produce fine-grained instance boundaries; 3) the RoI representation (proposal-based features) is redundant which impact the  inference speed with the increasing number of proposals. Compared to the existing methods, our method explicitly models spatial relation of instance, which helps us achieve high granularity for instance masks. We circumvent RoI operations for high-level tasks (a.k.a tracking and segmentation) by using collaborative connections, which link each task head interdependently for joint optimization and boost the computational tractability. \\ 
\indent \textbf{Data annotator.} MOTS data requires the instance mask as well as the temporal tracking label across frames.  So far, many attempts for semi-automated annotation~\cite{arun2020weakly,gupta2019lvis,xu2018srda,fang2019instaboost,khoreva2017simple,xu2016deep} have been made to reduce the annotation overhead. Aside from heavy engagement of human correction efforts~\cite{vondrick2013efficiently}, these methods generally have arduous implementations for the annotator and require multiple steps to achieve a desirable result. Moreover, their annotators cannot operate on a tracking level and only create instance masks in a single image. To produce the MOTS data, \cite{voigtlaender2019mots} leverages a refinement network to generate an initial segmentation mask followed by human corrections. The tracking labels are then created by delineating the temporal coherence across video frames~\cite{voigtlaender2019mots}. To our best knowledge, ~\cite{voigtlaender2019mots,porzi2020learning} are two methods available for  MOTS annotation, both built on Mask R-CNN by simply adding a tracking branch~\cite{he2017mask}. Compared to~\cite{voigtlaender2019mots,porzi2020learning}, our DG-Labeler explicitly models spatial relation to achieving fine-grained instance boundary. We also devise a collaborative connection, which uses the detection results to guide the high-level tasks (segmentation  and  tracking) to accurately fire on the task-relevant pixels. With limited human corrections, our annotation protocol can produce MOTS labels with appealing quality. 
\begin{figure*}[!ht]
		\centering
		\includegraphics[width=13cm]{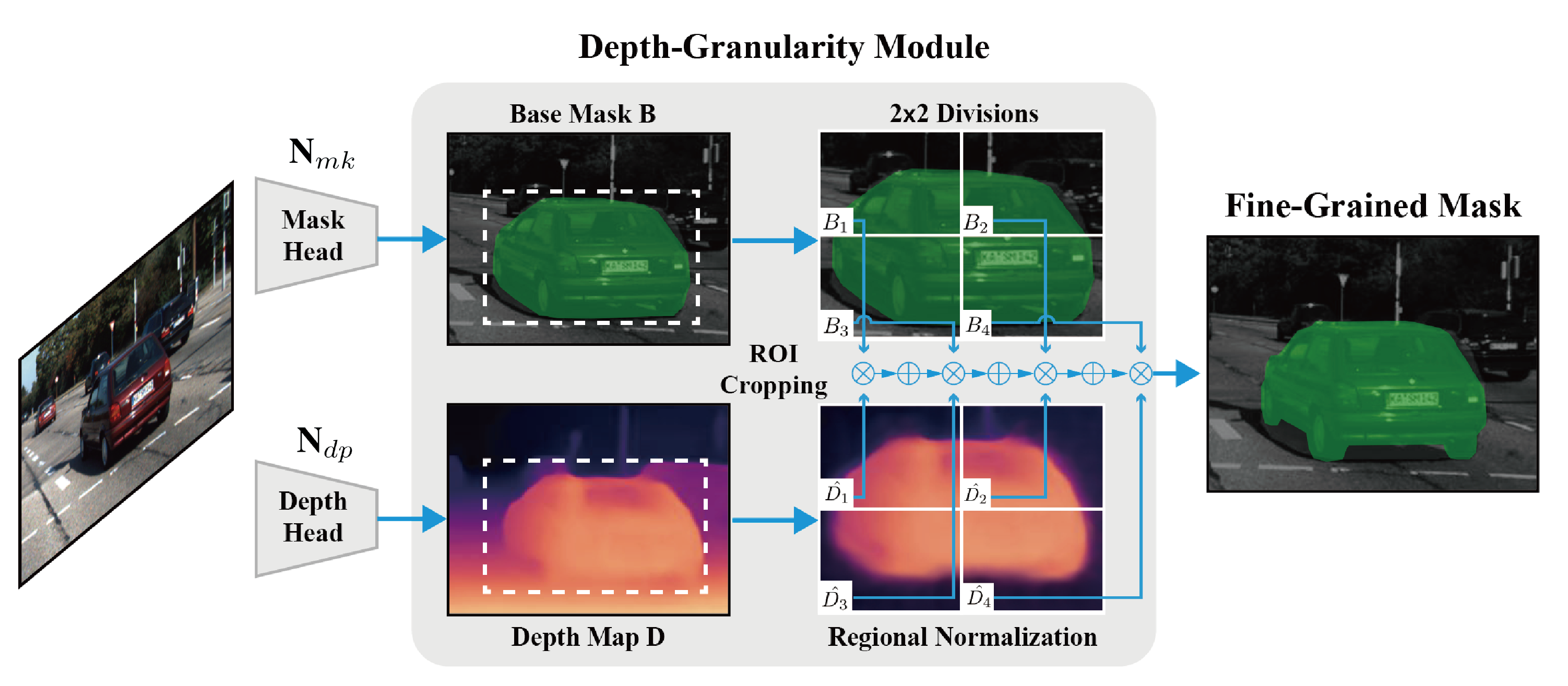}
		\caption{
		In the depth-granularity module,
		base mask and depth map are first divided into $2\times 2$ sub-regions. Then, each corresponding sub-region from the base mask and depth map are organically blended to produce the final fine-grained mask.}
		\label{DGM}
	\end{figure*}
\section{DG-Labeler}
\subsection{Overall Architecture}
Built on the  TrackR-CNN~\cite{voigtlaender2019mots}, our DG-Labeler includes architectures of feature extraction, task heads, and depth-granularity module, which collaboratively perform detection, segmentation, and tracking. Our overall architecture is shown in Figure~\ref{Frame}.\\
\indent \textbf{Feature extraction} uses the ResNet~\cite{he2016deep} backbone $\textbf{N}_{fm}$ to compute per frame feature maps $F$ and leverage a flow network $\textbf{N}_{fl}$ to model temporal features over time on video. Feature maps from the previous moment $F_{t-i}$ are warped into the current time $t$ based on the flow field $\Delta D_{t-i\rightarrow{t}}$ to obtain $F'_{t-i\rightarrow{t}}$. Afterwards, $F_{t}$ and $F'_{t-i\rightarrow{t}}$ are aggregated for $F'_{t}$ to increase the feature representation of the current frame. In default, the temporal range is 
three, namely, adjacent frames are used for the feature aggregation.\\
\indent \textbf{Task heads} are consisted of four task heads (a.k.a. the classification head $\textbf{N}_{cl}$, the bounding box head $\textbf{N}_{bb}$, the mask head $\textbf{N}_{mk}$, and the tracking head $\textbf{N}_{tr}$). The aforementioned task heads follow the implementation in TrackR-CNN. Besides, we craft a new depth head $\textbf{N}_{dp}$~\cite{godard2019digging} into our network to model the spatial relation of each detected object on the video frame. Since our feature extraction is built on ResNet~\cite{he2016deep}, we replace the U-Net architecture in~\cite{godard2019digging} with feature pyramid networks (FPNs)~\cite{lin2017feature} to predict depth maps. Other implementations are the same in ~\cite{godard2019digging}.\\
\indent \textbf{Depth-granularity module} is the key component of our DG-Labeler. The next section will detail this module.
\subsection{Depth-Granularity Module}
Inspired by~\cite{liu2021sg}, we blend the base mask $B$ with the corresponding depth map $D$ to generate the final fine-grained mask in the depth-granularity module (Figure~\ref{DGM}). In our implementation, both $B$ and $D$ have the same shape of $H\times W\times 1.$ We crop out each region of interest (RoI) on the base mask $B$ and the depth map $D$ based on the detected bounding boxes and divide each RoI into $k\times k$ regions of the same size. The division is arbitrary but we find that $k=2$ has the best speed-accuracy tradeoff. Afterward, each sub-region of the depth map is normalized by:
\begin{equation}
    \hat{D_{i}}=\frac{D_{i}-\min(D_{i})}{\max(D_{i})-\min(D_{i})}, \quad {\forall i\in\{1,...,k\}}, 
    \label{Equation 1}
\end{equation}
where $D_{i}$ and $\hat{D_{i}}$ represent one sub-region of the depth map and its normalized depth map respectively. $\hat{D_{i}}$  renders the spatial relation (foreground and background) and boundary details of the target instance (Figure~\ref{DGM}). \\
\indent Finally, we apply element-wise productions between the base mask and the normalized depth map from each corresponding sub-region, and sum along the $k\times k$
regions to obtain the final mask $M_j$ of the $j^{th}$ instance on the frame:
\begin{equation}
    M_j = \sum_{i=1}^{k\times k}\sigma(B_{i}\times  \hat{D_{i}}),
    \label{Equation 2}
\end{equation}
where $B_{i}$ and $\hat{D_{i}}$  are one sub-region of the base mask and the normalized depth map respectively, and $\sigma$ is \texttt{sigmoid} activation. In our implementation, our base mask $B_{i}$ uses floating point and the depth map $\hat{D_{i}}$ encodes the relative spatial relation, not the absolute depth values. With the spatial relation modeling, our final mask is more fine-grained.
\subsection{Collaborative Connections}
Unlike TrackR-CNN and its variants ~\cite{voigtlaender2019mots,luiten2020unovost,lin2019agss} whose task heads operate independently and ignore the intrinsic correlations among each task, we devise collaborative connections (the red lines in Figure~\ref{Frame}) across detection, segmentation, and tracking heads to facilitate the information prorogation across tasks. Compared to TrackR-CNN and its variants, this implementation offers us two improvements for the network behaviors: 1). our segmentation and tracking head fire on instance-centric features (the blue lines in Figure~\ref{Frame}) governed by the bounding boxes and the mask-guided boxes respectively, thus can perform more accurate predictions; 2) we improve the runtime performance by avoiding encoding redundant features based on  proposals produced by RPN and thus reducing computational cost per instance.\\
\subsection{Training Objective}
GIoU learning~\cite{rezatofighi2019generalized} is used in our training. Since our method follows the top-down paradigm, we argue that the improved bounding box regression can benefit the instance segmentation and tracking task. Bear this in mind, we leverage the \texttt{GIoU} loss in~\cite{liu2021sg} to organize our learning. Particularly, we propose a modified \texttt{GIoU} loss using a logarithmic function to increase the bounding box losses in order to facilitate hard  sample  learning ($i.e.,$ small GIoU):
\begin{equation}
\mathcal{L}_{box} =-\ln\frac{1+GIoU}{2}
\label{box}
\end{equation}
\indent Consequently, our overall loss can be defined as:
\begin{equation}
\mathcal{L}_{all} = \mathcal{L}_{box}+\mathcal{L}_{cls}+\mathcal{L}_{mask}+\mathcal{L}_{track}+\mathcal{L}_{depth}
\label{all}
\end{equation}
where $\mathcal{L}_{cls}, \mathcal{L}_{mask}$, and $\mathcal{L}_{track}$ are from \cite{voigtlaender2019mots},  $\mathcal{L}_{box}$ is the modified \texttt{GIoU} loss from Eq. \ref{box}, and $\mathcal{L}_{depth}$ is the average of per-pixel smoothness and masked photo- metric loss in \cite{godard2019digging}. Our architecture is trained in an end-to-end fashion.

\begin{table*}[!ht]
\centering
    \begin{tabular}{c|c|c|c|c|c}
    \toprule
    Datasets     & Video clip & Total frames & Identities & Instances & Ins./Fr. \\ \hline
    KITTI MOTS   & 21         & 8K           & 749        & 38K       & 4.78     \\ \hline
    BDD100K MOTS & 70         & 14K          & 6.3K       & 129K      & 9.20     \\ \hline
    Ours         & 40         & 12K          & 1.6K       & 68K       & 6.23     \\ 
    \bottomrule
    \end{tabular}
\caption{Annotation statistics. Our dataset outperforms the KITTI MOTS in annotation volume and density. BDD100K offers the largest training data but selected sequentially from video frames, which include redundant temporal information.}
\label{stats}
\end{table*}

\section{DGL-MOTS Dataset}
\subsection{Data Acquisition}
Our data acquisition is carefully designed to capture the high variability of driving scenarios, such as highway, local, residential, and parking. Our raw data is acquired from a moving vehicle with a span of two months, covering different lighting conditions in four different states in the USA. Images are recorded with a GoPro HERO8 at a frame rate of 17~\texttt{Hz}, behind the windshield of the vehicle. We deliberately skip post-processing ($i.e.,$ rectification or calibration) and keep data with motion blur and defocus to increase data diversity. We argue that data with a low-degree motion blur and defocus can better reflect the driving scenarios. However,  severely compromised video frames are excluded from annotations. 40 video sequences are manually selected for dense annotations, aiming for a high diversity of foreground objects (vehicles and pedestrians) and overall scene layout. Our annotation is elaborated in the next section.  
\subsection{Annotation Protocol}
To keep our annotation effort manageable, we use an iterative semi-automated annotation protocol based on our DG-Labeler. At the first iteration, we use the pre-trained DG-Labeler to automatically perform annotations for our data, followed by a manual correction step.  Per iteration, we fine-tune our DG-Labeler using the annotated data after manual corrections. We iterate the aforementioned process until pixel-level accuracy for all instances has been reached.\\ 
\indent To initialize DG-Labeler,
we use  ResNet-101~\cite{he2016deep} pre-trained on COCO~\cite{lin2014microsoft} and Mapillary~\cite{MVD2017} datasets as our feature extraction backbone; FlowNet pre-trained on the Flying Chair dataset~\cite{dosovitskiy2015flownet} is used to predict flow filed; and the depth network~\cite{godard2019digging} pre-trained on the KITTI depth dataset~\cite{uhrig2017sparsity} is used to predict depth map. Note training the depth network~\cite{godard2019digging} uses a self-supervised manner and only needs video sequences (3 consecutive frames without ground truth) to train. At the initial training, the weights of ResNet-101 and FlowNet are fixed, and the other weights related to different task heads are updated by learning on KITTI MOTS and BDD100K. We train the initial model for 40 epochs with a learning rate of $5 \times {10}^{-7}$ with Adam~\cite{kingma2014adam} optimizer and mini-batch size of 8. After each correction, the refined annotations are used to fine-tune our DG-Labeler.\\ 
\indent Eventually, we use 6 iterations to finalize the annotation process. \textit{We perform further processing on the annotated data and select learning examples for training and testing in every 5 frames. Following this design, our dataset generally has longer temporal representations and  descriptions.} Instead of splitting our annotated data randomly, we want to ensure that the training, validation, and test sets include the data representation for different driving scenarios, such as highway, local, residential, and parking areas (Figure 1). 

\begin{table*}[ht]
	\centering
		\scriptsize 
	\begin{tabular}{c|c|c|c|c|c|c|c|c|c|c|c}
		\toprule    
	\multirow{2}{*}{Method} & \multirow{2}{*}{Ep.} & Training&Testing & \multicolumn{4}{c|}{Cars} & \multicolumn{4}{c}{Pedestrians} \\
		\cline{5-12}
        \rule{0pt}{8pt}& & Dataset &Dataset & HOTA $\uparrow$ & sMOTSA$\uparrow$ & MOTSA$\uparrow$ & IDS$\downarrow$ & HOTA $\uparrow$ & sMOTSA$\uparrow$ & MOTSA$\uparrow$ &  IDS$\downarrow$\\
		\midrule
		PointTrack++  & 40 & KITTI & KITTI& $67.28$ & $82.82$ & $92.61$ & $\textcolor{blue}{\textbf{36}}$ & $56.67$ & $68.13$ & $83.67$ & $36$\\
		PointTrack++  & 40 & DGL-MOTS&KITTI & $\textcolor{red}{\textbf{68.40}}_{\textcolor{red}{\uparrow \textbf{1.12}}}$ & $\textcolor{red}{\textbf{84.33}}_{\textcolor{red}{\uparrow \textbf{1.51}}}$ & $\textcolor{red}{\textbf{93.68}}_{\textcolor{red}{\uparrow \textbf{1.07}}}$ &  $\textcolor{red}{\textbf{32}}_{\textcolor{red}{\downarrow \textbf{4}}}$ & $\textcolor{blue}{\textbf{57.87}}_{\textcolor{blue}{\uparrow \textbf{1.2}}}$ & $\textcolor{red}{\textbf{69.30}}_{\textcolor{red}{\uparrow \textbf{1.17}}}$ & $\textcolor{red}{\textbf{84.51}}_{\textcolor{red}{\uparrow \textbf{0.84}}}$ & $\textcolor{red}{\textbf{33}}_{\textcolor{red}{\downarrow \textbf{3}}}$\\
	PointTrack++ & 20 & DGL-MOTS &KITTI& $\textcolor{blue}{\textbf{67.42}}_{\uparrow \textcolor{blue}{\textbf{0.14}}}$ & $\textcolor{blue}{\textbf{82.99}}_{\uparrow \textcolor{blue}{\textbf{0.6}}}$ & $\textcolor{blue}{\textbf{92.67}}_{\uparrow \textcolor{blue}{\textbf{0.6}}}$ &  $\textcolor{blue}{\textbf{36}}_{\textcolor{blue}{\downarrow \textbf{0}}}$ & $\textcolor{red}{\textbf{57.89}}_{\textcolor{red}{\uparrow \textbf{1.22}}}$ & $\textcolor{blue}{\textbf{68.20}}_{\textcolor{blue}{\uparrow \textbf{0.07}}}$ & $\textcolor{blue}{\textbf{83.76}}_{\textcolor{blue}{\uparrow \textbf{0.09}}}$ & $\textcolor{blue}{\textbf{35}}_{\downarrow \textcolor{blue}{\textbf{1}}}$\\
		\midrule
	TrackRCNN & 40 & KITTI & KITTI & $56.92$ & $77.20$ & $87.92$ & $92$ & $42.08$ & $47.43$ & $67.12$  & $78$ \\
TrackRCNN & 40 & DGL-MOTS& KITTI & $58.02_{\uparrow 1.1}$ & $78.80_{\uparrow 1.60}$ & $88.90_{\uparrow 1.00}$& $80_{\downarrow 12}$  & $43.11_{\uparrow 1.03}$ & $48.61_{\uparrow 1.18}$  & $68.33_{\uparrow 1.21}$ & $62_{\downarrow 16}$\\
		TrackRCNN& 20 & DGL-MOTS& KITTI & $57.12_{\uparrow 0.20}$ & $77.31_{\uparrow 0.11}$ & $88.15_{\uparrow 0.23}$ & $90_{\downarrow 2}$ & $42.21_{\uparrow 0.13}$ & $47.50_{\uparrow 0.07}$  & $68.46_{\uparrow 1.34}$ & $76_{\downarrow 2}$\\
			\midrule
		STEm-Seg & 40 & KITTI& KITTI & $56.36$ & $76.30$ & $86.63$ & $76$ & $43.10$ & $51.02$ & $66.60$ & $74$\\
		STEm-Seg & 40 & DGL-MOTS&KITTI & $57.50_{\uparrow 1.14}$ & $77.35_{\uparrow 1.05}$ & $87.92_{\uparrow 1.29}$ & $56_{\downarrow 20}$ & $45.10_{\uparrow2.00}$ & $52.70_{\uparrow 1.68}$ & $68.00_{\uparrow 1.4}$& $60_{\downarrow 14}$\\
		STEm-Seg & 20 & DGL-MOTS&KITTI & $56.70_{\uparrow 0.34}$ & $76.36_{\uparrow 0.06}$ & $86.70_{\uparrow 0.07}$ & $66_{\downarrow 10}$ & $43.45_{\uparrow 0.35}$ & $51.42_{\uparrow 0.4}$ & $66.99_{\uparrow 0.39}$& $70_{\downarrow 4}$\\
			\midrule
	PointTrack++ & 40 &BDD100K&BDD100K& $\textcolor{blue}{\textbf{68.33}}$ & $\textcolor{blue}{\textbf{84.60}}$ & $93.20$ &  $\textcolor{blue}{\textbf{49}}$ & $55.42$ & $\textcolor{blue}{\textbf{64.56}}$ & $\textcolor{blue}{\textbf{80.29}}$ & $45$  \\
		PointTrack++  & 40 & DGL-MOTS&BDD100K & $\textcolor{red}{\textbf{69.28}}_{\textcolor{red}{\uparrow \textbf{0.95}}}$ & $\textcolor{red}{\textbf{85.59}}_{\textcolor{red}{\uparrow \textbf{0.99}}}$ & $\textcolor{red}{\textbf{94.32}}_{\textcolor{red}{\uparrow \textbf{1.12}}}$ &  $\textcolor{red}{\textbf{38}}_{ \textcolor{red}{\downarrow\textbf{11}}}$ & $\textcolor{red}{\textbf{56.89}}_{\textcolor{red}{\uparrow \textbf{1.47}}}$ & $\textcolor{red}{\textbf{65.28}}_{\textcolor{red}{\uparrow \textbf{0.72}}}$ & $\textcolor{red}{\textbf{81.05}}_{\textcolor{red}{\uparrow \textbf{0.76}}}$ & $\textcolor{red}{\textbf{34}}_{\textcolor{red}{\downarrow \textbf{11}}}$\\
	PointTrack++ & 20 & DGL-MOTS &BDD100K& $68.26_{{\downarrow 0.07}}$ & $84.43_{{\downarrow 0.17 }}$ & $\textcolor{blue}{\textbf{93.27}}_{\textcolor{blue}{\uparrow \textbf{0.07}}}$ &  $52_{{\uparrow 3}}$ & $\textcolor{blue}{\textbf{55.37}}_{{\downarrow \textcolor{blue}{\textbf{0.05}}} }$ & $64.23_{{\downarrow 0.33 }}$ & $80.26_{{\downarrow 0.03}}$ & $50_{{\uparrow 5}}$\\
		\midrule
	TrackRCNN & 40 &BDD100K&BDD100K & $57.91$ & $78.10$ & $88.62$ &  $85$ & 46.37 & $55.93$ &  $70.18$& $88$\\
TrackRCNN & 40 & DGL-MOTS& BDD100K & $59.22_{\uparrow 1.31}$ & $79.82_{\uparrow 1.72}$ & $89.90_{\uparrow 1.08}$& $68_{\downarrow 17}$ & $47.49_{\uparrow 1.42}$ & $56.61_{\uparrow 0.68}$  & $71.80_{\uparrow 1.62}$ & $78_{\downarrow 10 }$\\
		TrackRCNN& 20 & DGL-MOTS& BDD100K & $58.09_{\uparrow 0.18}$ & $78.20_{\uparrow 0.10}$ & $88.69_{\uparrow 0.07}$ & $80_{\downarrow 5}$ & $46.52_{\uparrow 0.15}$ & $56.07_{\uparrow 0.14}$  & $70.32_{\uparrow 0.14}$& $84_{\downarrow 4}$\\
    	\midrule
		STEm-Seg & 40 &BDD100K  &BDD100K& $57.39$ & $77.24$ &$87.65$ & $66$ & $47.65$ & $56.30$ & $71.03$ & $48$\\
		STEm-Seg & 40 & DGL-MOTS&BDD100K & $58.62_{\uparrow 1.23}$ & $78.50_{\uparrow 1.20}$ & $88.96_{\uparrow 1.30}$ & $56_{\downarrow 10}$ & $49.00_{\uparrow 1.35}$ & $57.72_{\uparrow 1.42}$ & $72.20_{\uparrow 1.17}$ & $\textcolor{blue}{\textbf{38}}_{\downarrow \textcolor{blue}{\textbf{10}}}$\\
		STEm-Seg & 20 & DGL-MOTS&BDD100K & $57.70_{\uparrow 0.31}$ & $77.78_{\uparrow 0.54}$ & $88.04_{\uparrow 0.39}$ & $64_{\downarrow 2}$ & $47.95_{\uparrow 0.30}$ & $56.98_{\uparrow 0.68}$ & $71.50_{\uparrow 0.47}$ & $44_{\downarrow 4}$\\
		\bottomrule
	\end{tabular}
\caption{The results for cross-dataset evaluation on KITTI, BDD100K, and our DGL-MOTS.
${\uparrow}$ and ${\downarrow}$ indicate the change of performance on the metrics. The \textcolor{red}{best} and the \textcolor{blue}{second-best} methods on KITTI and BDD100K are highlighted.}
		\label{trainS}
	\end{table*}
\begin{figure}[h]
	\centering
	\includegraphics[width=8.5cm]{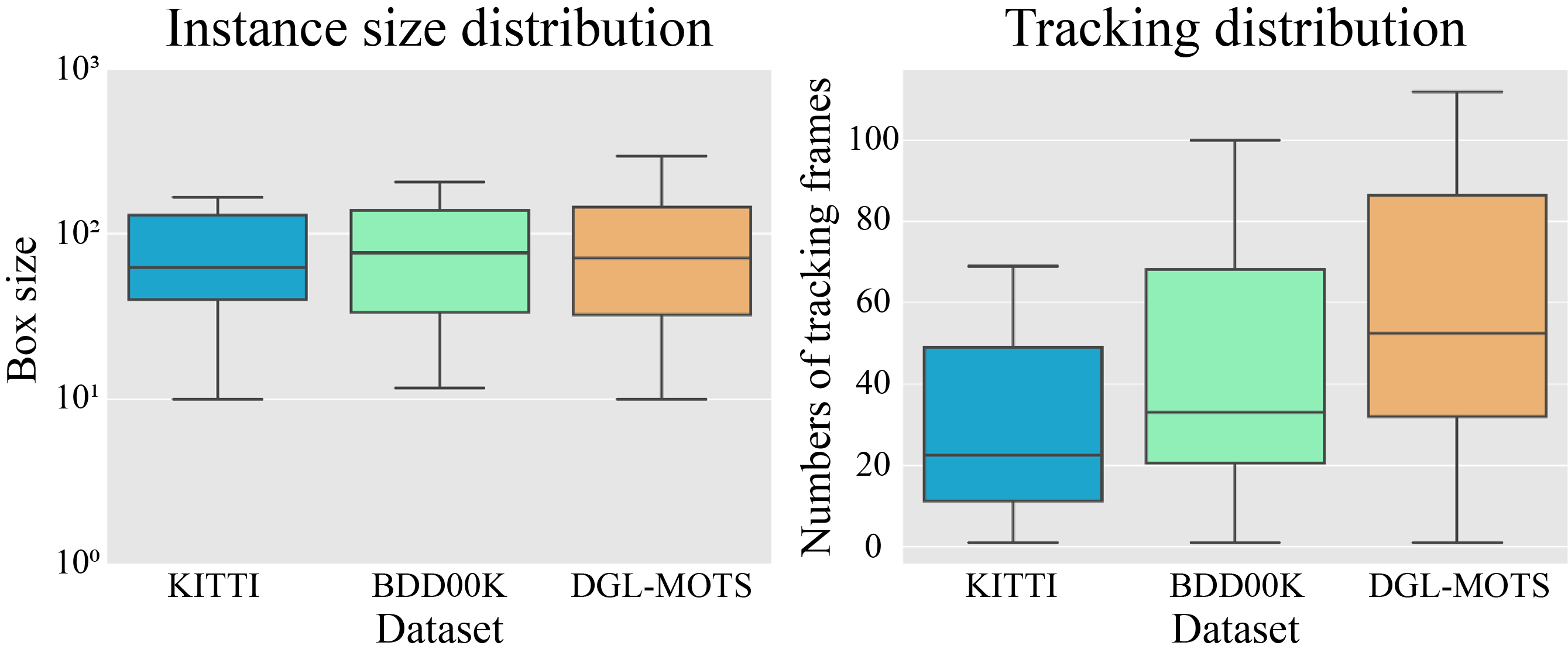}
	\caption{Distribution of the instance size (left) based on the bounding box size and track length (right) based on the duration of instances that appeared on video. Our dataset is more diverse in object scale and tracking length than counterparts. 
	}
	\label{DStats}
\end{figure}
\section{Experiments}
\subsection{Implementation Details}
In order to evaluate the proposed dataset and annotator, we perform cross-dataset evaluations and compare against three recent state-of-the-art MOTS methods\footnote{All the compared methods use ResNet101~\cite{he2016deep}.} (a.k.a. PointTrack++~\cite{xu2020segment}, TrackRCNN~\cite{voigtlaender2019mots}, and STEm-Seg~\cite{athar2020stem}, which only require an instance-level label for training \footnote{There are more progressive methods \cite{qiao2020vip,kimeagermot,luiten2020track}. However, these methods needs extra information (i.e., flow field and LiDAR measurement) for supervision.}. All the methods are trained on the KITTI MOTS, BDD100K, and DGL-MOTS train sets separately and cross-validated on each dataset.
In training, we designate no fixed number of total iterations and allow each method to be trained until performance asymptotically. The evaluation metrics are sMOTSA, MOTSA, IDS and HOTA from~\cite{voigtlaender2019mots}. All experiments are conducted on one TITAN RTX GPU.
\subsection{Dataset Statistics}
\textbf{Annotation volume} is summarized in Table~\ref{stats}. We compare  DGL-MOTS with BDD100K and KITTI MOTS in terms of the number of video clips, video frames, unique identities, instances, and instances per frame. In comparison, DGL-MOTS outperforms KITTI MOTS in evaluation metrics. Particularly, our instances per frame are around 1.5\% higher than that of KITTI MOTS, which indicates that our dataset has a higher portion of scene complexity. BDD100K has the largest data volume among the three datasets, but its data is intensively selected from sequential video frames, which include redundant learning examples.\\
\indent \textbf{Instance variations} are represented by the instance appearance change as well as the temporal description (as shown in Figure \ref{DStats}). The left figure illuminates the distribution of squared bounding-box size $\sqrt{wh}$ (where width $w$ and height $h$); while the right figure shows the distribution  of tracking length per instance. Figure \ref{DStats} demonstrates that our dataset is not only more diverse in visual scale, but also longer in the temporal range for tracking.\\ 
\indent \textbf{Scene diversity} is well-represented in our DGL-MOTS dataset, which includes more diverse driving scenes (Figure 1). Since DGL-MOTS provides recordings from four different states, it covers significantly more areas than KITTI MOTS that contains driving footage from a single city (Karlsruhe, German). Compared to BDD100K, our dataset include more road settings such as parking, residential, local, and high-way, while BDD100K only collects data of inner-city from the populous areas in the US \cite{yu2020bdd100k}.
\begin{figure*}[!ht]
	\centering
	\includegraphics[width=16.6cm]{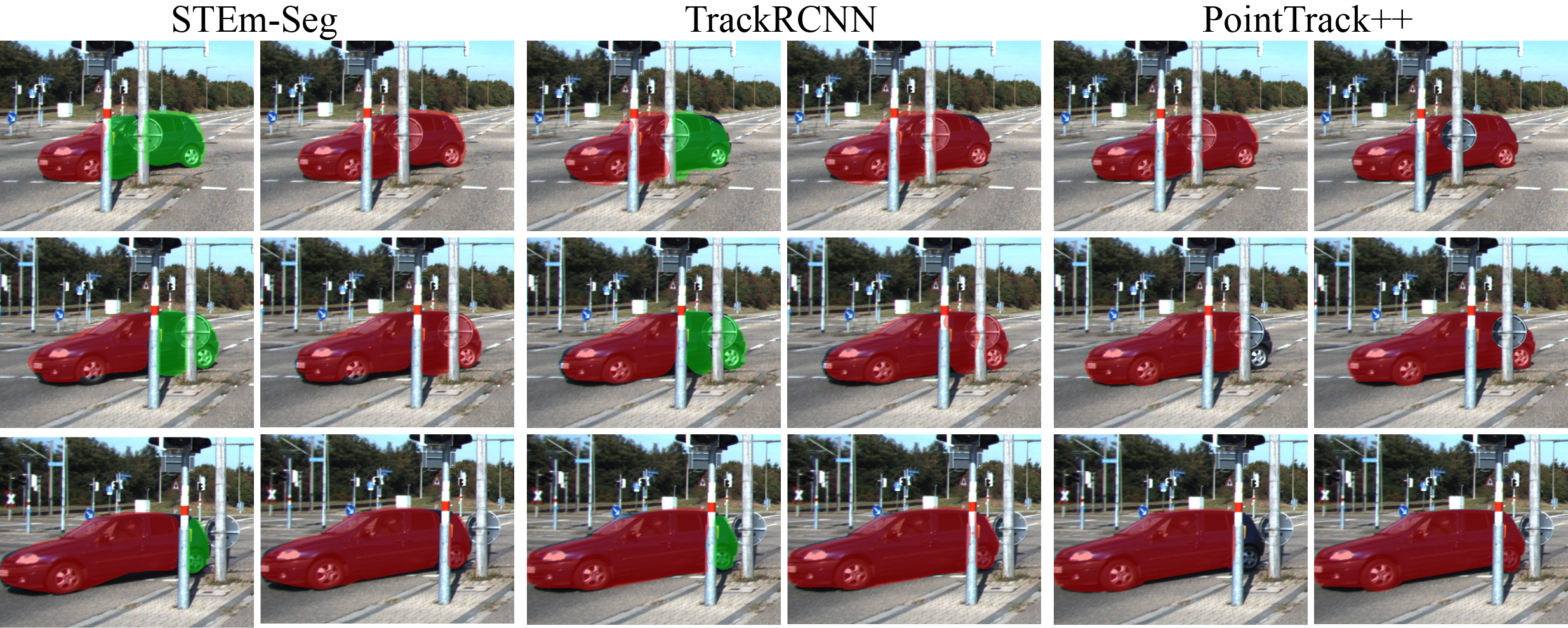}
	\caption{Qualitative examples of different methods tested on the KITTI test set. Results on the left column are methods trained on the KITTI while results on the right column are methods trained on our dataset. We can see the improvement of segmentation and tracking using our dataset.
	Masks of the same color indicate the tracking of the same instance.}
	\label{cdt}
\end{figure*}
		\begin{table*}[ht]
	\centering
		\scriptsize
	\begin{tabular}{c|c|c|c|c|c|c|c|c|c}
		\toprule    
		\multirow{2}{*}{Method} & \multirow{2}{*}{Dataset} &  \multicolumn{4}{c|}{Cars} & \multicolumn{4}{c}{Pedestrians} \\
		\cline{3-10}
        \rule{0pt}{12pt} & & HOTA$\uparrow$ & sMOTSA$\uparrow$ & MOTSA$\uparrow$ & IDS$\downarrow$ &  HOTA$\uparrow$ &sMOTSA$\uparrow$ & MOTSA$\uparrow$ & IDS$\downarrow$ \\
					\midrule
		PointTrack++  & KITTI & \textcolor{blue}{\textbf{67.28}} & \textcolor{blue}{\textbf{82.82}} & \textcolor{red}{\textbf{92.61}} & \textcolor{blue}{\textbf{36}} & \textcolor{red}{\textbf{56.67}} & \textcolor{blue}{\textbf{68.13}} & \textcolor{red}{\textbf{83.67}} & \textcolor{red}{\textbf{36}}\\
		TrackRCNN & KITTI  & 56.92 & 77.20 & 87.92 & 92 & 42.08 & 47.43 &67.12 & 78\\
		STEm-Seg & KITTI  & 56.36 & 76.30 &86.63 & 76 & 43.10 & 51.02 & 66.60&74\\
		\textbf{DG-Labler (Ours)} & KITTI  & $\textcolor{red}{\textbf{69.72}}$ & $\textcolor{red}{\textbf{83.68}}$ & $\textcolor{blue}{\textbf{90.72}}$ & $\textcolor{red}{\textbf{35}}$ & $\textcolor{blue}{\textbf{55.90}}$ & $\textcolor{red}{\textbf{69.36}}$ & $\textcolor{blue}{\textbf{83.40}}$ & $\textcolor{blue}{\textbf{50}}$\\
		\midrule
			PointTrack++  & BDD100K & \textcolor{red}{\textbf{68.33}} & \textcolor{blue}{\textbf{84.60}} & \textcolor{red}{\textbf{93.20}} & \textcolor{red}{\textbf{49}} & \textcolor{blue}{\textbf{55.42}} & \textcolor{blue}{\textbf{64.56}} & \textcolor{blue}{\textbf{80.29}}& \textcolor{red}{\textbf{45}}\\
		TrackRCNN & BDD100K & 57.91 & 78.10 & 88.62 & 85 & 46.37 & 55.93 &70.18 & 88\\
		STEm-Seg & BDD100K & 57.39 & 77.24 &87.65 & 60 & 47.65 & 56.30 & 71.03 & 48\\
		\textbf{DG-Labler (Ours)} & BDD100K  & $\textcolor{blue}{\textbf{67.89}}$ & $\textcolor{red}{\textbf{85.30}}$ & $\textcolor{blue}{\textbf{91.70}}$ & $\textcolor{blue}{\textbf{58}}$ & $\textcolor{red}{\textbf{56.23}}$ & $\textcolor{red}{\textbf{65.43}}$ & $\textcolor{red}{\textbf{81.40}}$ & $\textcolor{blue}{\textbf{48}}$\\
		\midrule
		PointTrack++ & DGL-MOTS & \textcolor{blue}{\textbf{68.10}} & \textcolor{blue}{\textbf{83.62}} & \textcolor{red}{\textbf{92.39}} & \textcolor{blue}{\textbf{42}} & \textcolor{blue}{\textbf{59.10}} & \textcolor{blue}{\textbf{71.90}} & \textcolor{blue}{\textbf{86.60}} & \textcolor{blue}{\textbf{32}}\\
		TrackRCNN & DGL-MOTS  & 58.63 & 78.8 & 88.9 & 88 & 48.23 & 60.29 & 76.10 & 77\\
		STEm-Seg & DGL-MOTS & 57.90 & 77.99 &87.9 & 78 & 47.88 & 59.82 & 67.70 & 58\\
		\textbf{DG-Labler (Ours)} & DGL-MOTS & \textcolor{red}{\textbf{69.35}} & \textcolor{red}{\textbf{84.10}} & \textcolor{blue}{\textbf{91.43}} & \textcolor{red}{\textbf{40}} & \textcolor{red}{\textbf{61.20}} & \textcolor{red}{\textbf{73.17}} & \textcolor{red}{\textbf{87.14}} & \textcolor{red}{\textbf{28}}\\
		\bottomrule
	\end{tabular}
\caption{Comparison with the state-of-the-art methods on the KITTI MOTS, BDD100K, and DGL-MOTS. Each method is trained on KITTI MOTS, BDD100K, and  DGL-MOTS separately. The \textcolor{red}{best} and the \textcolor{blue}{second-best} methods are highlighted.} 
		\label{trainD}
	\end{table*}
\subsection{Cross-Dataset Evaluations}
Table~\ref{trainS} reports the results for the cross-dataset evaluations to assess our DGL-MOTS dataset. For the same method trained on different datasets, their performance gaps stem from the quality of the dataset ($i.e.,$ annotation quality, data diversity, and temporal representation). Essentially, we observe two benefits of using our dataset in training over its counterparts. First,  methods trained on DGL-MOTS all outperform their counterparts (with the same network architecture) on all metrics (Table~\ref{trainS}). The improved performance indicates that, compared to KITTI and BDD100K, our dataset captures more general road settings and driving scenarios in training. Second, the DGL-MOTS dataset can train methods to achieve improved performance with a shorter schedule than methods trained on KITTI and BDD100K. For instance, TrackRCNN~\cite{voigtlaender2019mots} and STEm-Seg~\cite{athar2020stem} trained on DGL-MOTS with 20 epochs outperform its counterpart trained on KITTI and BDD100K with 40 epochs respectively.\\ 
\indent In addition, we display the qualitative examples of each method from KITTI MOTS in Figure ~\ref{cdt}. The selected results also resonate with  our quantitative analysis that methods trained on our DGL-MOTS dataset generally achieve improved performance in instance mask generation and tracking than  their counterparts (with the same architecture) trained on KITTI MOTS. Both quantitative and qualitative results prove the advantages of the proposed DGL-MOTS dataset over the concurrent datasets. 

\begin{figure*}[!ht]
    \centering
    \includegraphics[width=16.2cm]{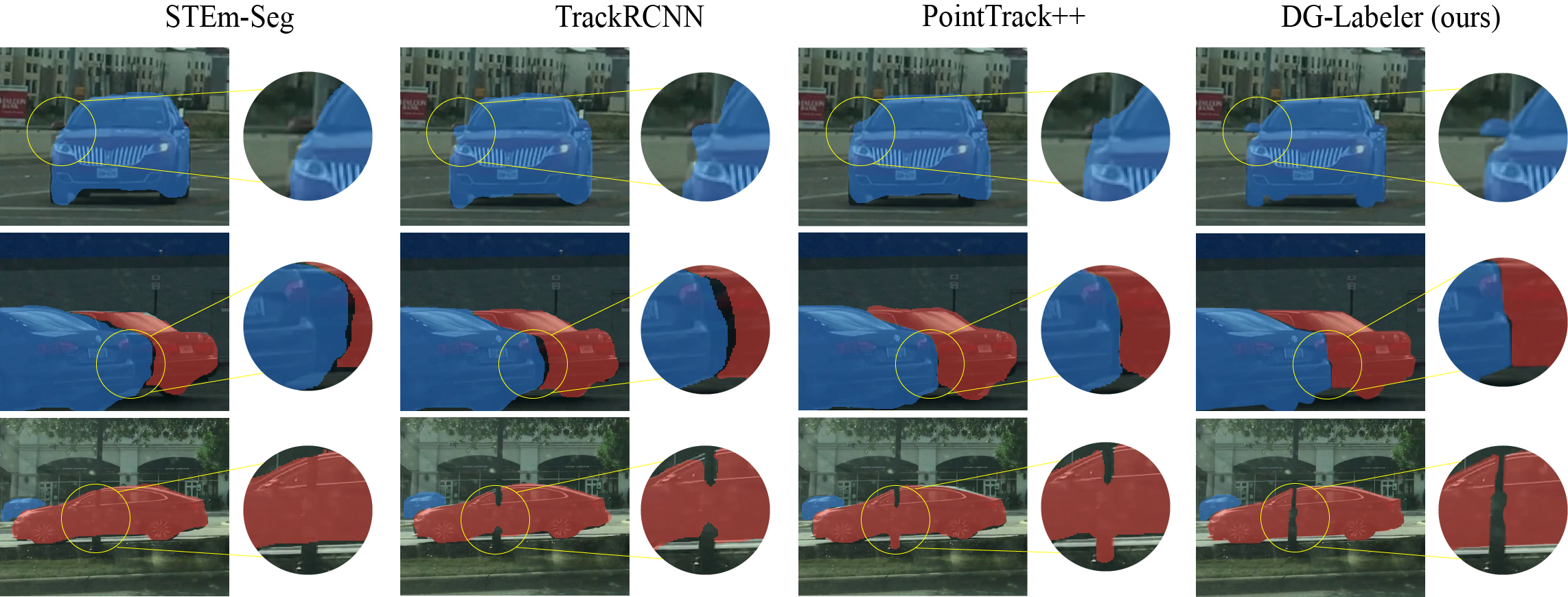}
    \caption{Qualitative examples of different methods on DGL-MOTS dataset. Compared to other methods, our DG-Labeler offers fine-grained instance masks. All methods are trained and tested on our DGL-MOTS dataset.}
    \label{qe}
\end{figure*}	
	\begin{table*}[ht]
	\centering
	\scriptsize
	\begin{tabular}{c|c|c|c|c|c|c|c|c}
		\toprule    
			\multirow{2}{*}{Method} & \multicolumn{4}{c|}{Cars} & \multicolumn{4}{c}{Pedestrians} \\
		\cline{2-9}
		\rule{0pt}{12pt} & HOTA$\uparrow$ & sMOTSA$\uparrow$ & MOTSA$\uparrow$ & IDS$\downarrow$ & HOTA$\uparrow$ & sMOTSA$\uparrow$ & MOTSA$\uparrow$ & IDS$\downarrow$\\
		\midrule
		TrackRCNN & 57.91 & 78.10 & 88.62 & 85 & 46.37 & 55.93 & 70.18 & 88\\
		TrackRCNN+CC & $63.51_{\uparrow 5.6}$ & $79.9_{\uparrow 1.8}$ & $89.42_{\uparrow 0.8}$ & $67_{\downarrow 18}$ & $48.77_{\uparrow 2.4}$ & $58.73_{\uparrow 2.8}$ & $72.18_{\uparrow 2.0}$ & $62_{\downarrow 26}$\\
		TrackRCNN+DGM & $66.81_{\uparrow 8.9}$ & $82.7_{\uparrow 4.6}$ & $90.92_{\uparrow 2.3}$ & $74_{\downarrow 11}$ & $53.07_{\uparrow 6.7}$ & $62.03_{\uparrow 6.1}$ & $77.48_{\uparrow 7.3}$ & $74_{\downarrow 14}$\\
		TrackRCNN+CC+GL & $64.51_{\uparrow 6.6}$ & $80.90_{\uparrow 2.8}$ & $90.92_{\uparrow 2.3}$ & $65_{\downarrow 20}$ & $50.27_{\uparrow 3.9}$ & $59.93_{\uparrow 4.0}$ & $73.78_{\uparrow 3.6}$ & $\textcolor{blue}{\textbf{55}}_{\textcolor{blue}{\downarrow \textbf{33}}}$\\
		TrackRCNN+CC+DGM & $\textcolor{blue}{\textbf{68.51}}_{\textcolor{blue}{\uparrow \textbf{10.6}}}$ & $\textcolor{blue}{\textbf{84.68}}_{\textcolor{blue}{\uparrow \textbf{6.4}}}$ & $\textcolor{blue}{\textbf{91.42}}_{\textcolor{blue}{\uparrow \textbf{2.8}}}$ & $\textcolor{blue}{\textbf{63}}_{\textcolor{blue}{\downarrow \textbf{22}}}$ & $\textcolor{blue}{\textbf{54.77}}_{\textcolor{blue}{\uparrow \textbf{8.4}}}$ & $\textcolor{blue}{\textbf{64.83}}_{\textcolor{blue}{\uparrow \textbf{8.9}}}$ & $\textcolor{blue}{\textbf{80.08}}_{\textcolor{blue}{\uparrow \textbf{9.9}}}$ & $60_{\downarrow 28}$\\
		\midrule
		\textbf{DG-Labeler (Ours)} & 
		$\textcolor{red}{\textbf{69.35}}_{\textcolor{red}{\uparrow \textbf{11.44}}}$ & $\textcolor{red}{\textbf{85.30}}_{\textcolor{red}{\uparrow \textbf{7.2}}}$ & $\textcolor{red}{\textbf{91.70}}_{\textcolor{red}{\uparrow \textbf{3.08}}}$ & $\textcolor{red}{\textbf{58}}_{\textcolor{red}{\downarrow \textbf{27}}}$ & $\textcolor{red}{\textbf{56.23}}_{\textcolor{red}{\uparrow \textbf{9.86}}}$ & $\textcolor{red}{\textbf{65.43}}_{\textcolor{red}{\uparrow \textbf{9.5}}}$ & $\textcolor{red}{\textbf{81.40}}_{\textcolor{red}{\uparrow \textbf{11.22}}}$ & $\textcolor{red}{\textbf{48}}_{\textcolor{red}{\downarrow \textbf{40}}}$\\
		\bottomrule
	\end{tabular}
		\caption{Ablation study results on the BDD100K. All methods are trained on the BDD100K training set. CC, DGM, and GL stand for collaborative connections, depth-granularity module, and GIoU loss respectively.  We use the best models in training for testing. ${\downarrow}$ and ${\uparrow}$ indicate the performance gain to the baseline. The \textcolor{red}{best} and the \textcolor{blue}{second-best} methods are highlighted.}
		\label{ab}
	\end{table*}
\subsection{Comparison to The State-Of-The-Art}
This section presents the state-of-the-art comparison of our DG-Labeler on KITTI, BDD100K, and DGL-MOTS.\\
\indent \textbf{Quantitative results} illuminates that DG-Labeler achieves the appealing performance on all metrics (on HOTA, sMOTSA, MOTSA, and IDS) among all methods. (Table~\ref{trainD}). Also, DG-Labeler is on-par with PointTrack++~\cite{xu2020segment}, the concurrent method. For instance, our margins over the strong methods (TrackRCNN~\cite{voigtlaender2019mots} and STEm-Seg~\cite{athar2020stem}) are  around 3.53-13.36\%  for  the car class  and  9.5-21.93\% for the pedestrian class on HOTA, sMOTSA, and MOTSA. The improvements suggest that DG-Labeler has a superior segmentation behavior to other recent methods. Meanwhile, DG-Labeler performs on par with the top-performing method, PointTrack++~\cite{xu2020segment} on all metrics. The reported results indicate that our DG-Labeler is competitive with the existing best approaches.\\
\indent \textbf{Qualitative examples} demonstrate the improved instance mask quality of our DG-Labeler over the counterpart methods (as shown in Figure~\ref{qe}). To demonstrate our advantage, we select some samples where other methods have trouble dealing with. Those cases include 1) objects with complex shapes ($i.e.,$ wing mirrors or pedestrians), which is hard to depict sharp borders; 2) same class objects with overlapping. Other methods often get confused with the borders and fail to segment accurate boundaries; 3) Objects in separated parts ($i.e.,$ occluded or truncated objects). Other methods may segment targets into separate objects or include occlusions as false positives. Based on the results, our DG-Labeler achieves an improved segmentation behavior in these cases because our depth-granularity module models the object spatial relations which offer more accurate descriptions of instance details and boundaries.
Besides, our collaborative connections allow our segmentation and tracking head to accurately fire on the pixel of the instance instead of using the candidate proposals.
\subsection{Ablation  Study}
We perform an ablation study on the  BDD100K test set. Note our method is crafted on TrackRCNN~\cite{voigtlaender2019mots}, thus our baseline. By progressively integrating different contributing components: collaborative connections (CC),  depth-granularity module (DGM), and GIoU loss (GL) (Sec. 3.4), to the baseline, we assess the contribution of each new component in DG-Labeler to TrackRCNN~\cite{voigtlaender2019mots}.\\ \indent We present the results in Table~\ref{ab}. All of our components (CC, DGM, GL) assist in achieving improved performance. Particularly for a single module, the baseline with CC avoids inefficient proposal-based operations and performs predictions on the accurate RoIs, thus achieving improved performance in accuracy; DGM contributes the largest improvements in dense prediction (HOTA, sMOTSA, and MOTSA). Compared to the strong baseline TrackRCNN, our full model integrating all contributions obtains absolute gains of 11.44\%, 7.2\%, 3.08\%, and 27 in terms of HOTA, sMOTSA, MOTSA, and IDS for car class and 9.86\%, 9.5\%, 11.22\%, and 40 for pedestrian class respectively. More results are displayed in the supplementary materials.
\section{Conclusion}
In this work, we offer the DGL-MOTS Dataset for training MOTS algorithm as well as DG-Labeler for data annotation. 
We believe that our work holds valuable potentials to facilitate the progress of the MOTS studies.


{\small
\bibliographystyle{ieee_fullname}
\bibliography{egbib}
}

\end{document}